\documentclass[sigconf]{acmart}

\copyrightyear{2026}
\acmYear{2026}
\acmConference[SIGSPATIAL '26]{The 34th ACM International Conference on
  Advances in Geographic Information Systems}{November 03--06,
  2026}{Riverside, CA, USA}
\acmBooktitle{The 34th ACM International Conference on Advances in
  Geographic Information Systems (SIGSPATIAL '26), November 03--06, 2026,
  Riverside, CA, USA}
\acmDOI{10.1145/3841645.3843391}
\acmISBN{979-8-4007-2950-8/2026/11}

\setcopyright{none}
\renewcommand\footnotetextcopyrightpermission[1]{}
\usepackage{booktabs}

\begin{document}

\title{Do Location Encoders Capture Spatial Effects? A GeoShapley Benchmark Across Scales}

\author{Daniel Kiv}
\affiliation{%
  \institution{Siebel School of Computing \& Data Science, University of Illinois Urbana-Champaign}
  \city{Urbana}
  \state{IL}
  \country{USA}}
\email{dkiv2@illinois.edu}

\author{Shaowen Wang}
\affiliation{%
  \institution{Department of Geography \& GIS and Siebel School of Computing \& Data Science, University of Illinois Urbana-Champaign}
  \city{Urbana}
  \state{IL}
  \country{USA}}
\email{shaowen@illinois.edu}

\renewcommand{\shortauthors}{Kiv and Wang}

\begin{abstract}
Location encoders transform geographic coordinates into high-dimensional
embeddings for downstream machine learning, but it is unclear how well these
representations capture interpretable spatial effects. We benchmark whether
GeoShapley, a game-theoretic explainer that treats all location features as a
single joint player, can recover spatially varying coefficients from models built
on location-encoder embeddings. Eleven encoders from the TorchSpatial framework
are evaluated against a synthetic process with known coefficients, across three
scales (grid, county, global), with and without raw coordinates alongside the
embedding, and under untrained and contrastively trained conditions. Measuring
recovery as the correlation between estimated and true coefficients, we report
how it varies with scale and encoder architecture. Recovery of the primary
coefficient is consistently high across encoders, whereas recovery of a
secondary coefficient is more scale-dependent, differing most at the global
scale; a raw-coordinate baseline remains competitive throughout.
\end{abstract}

\ccsdesc[500]{Computing methodologies~Supervised learning by regression}
\ccsdesc[500]{Information systems~Geographic information systems}
\ccsdesc[300]{Computing methodologies~Learning latent representations}
\ccsdesc[100]{General and reference~Evaluation}

\keywords{Location encoding, GeoAI, explainable AI, GeoShapley, spatially
varying coefficients, spatial interpretability}

\maketitle
\pagestyle{plain}

\section{Introduction}
\label{sec:introduction}

Spatial heterogeneity is a fundamental property of geographic processes:
relationships between variables change from place to place, so a single global
model often fails to explain geographic data well. Spatially varying
coefficient (SVC) models, such as Geographically Weighted Regression (GWR) and
its multiscale extension, capture this by letting coefficients change with
location \citep{fotheringhamMultiscaleGeographicallyWeighted2017,
anselinSpatialEconometricsMethods1988}. GWR is itself a form of local
statistical learning, and machine learning methods such as geographical random
forests \citep{georganosGeographicalRandomForests2021} likewise estimate
geographically heterogeneous relationships, so the distinction that matters here
is not statistics versus machine learning but how location enters the model:
through an explicit spatial kernel and a local linear form, or as features a
flexible model is free to use nonlinearly.

A common way to bring location into a machine learning model is a
\emph{location encoder}, a deep network that turns raw coordinates into
a high-dimensional embedding for a downstream task
\citep{maiReviewLocationEncoding2022, aodhaPresenceOnlyGeographicalPriors2019,
maiSphere2VecGeneralpurposeLocation2023}. Raw latitude and longitude are awkward
inputs: they are angular measures on a sphere, so a model need not recognize
that two points on opposite sides of the antimeridian are close together. A
location encoder addresses this with a hand-designed coordinate transform
followed by a neural network, which makes location a black box inside an already
black-box model. Geographic information is spread across all $D$ embedding
dimensions, none of which corresponds to an interpretable geographic quantity,
so feature-wise post-hoc explainers that score one feature at a time never
recover a single, coherent location effect.

GeoShapley \citep{liGeoShapleyGameTheory2024} addresses this by treating all
location features as a single joint player in the Shapley value framework
\citep{harrisJointShapleyValues2022}, collapsing the entire location
representation into one intrinsic location effect from which spatially varying
coefficients can be recovered. \citet{liGeoShapleyGameTheory2024} notes that an
embedding could be grouped into this player exactly as raw coordinates are, but
evaluates GeoShapley only on coordinates and analytic spatial bases
\citep{liExtractingSpatialEffects2022, liuEnsembleFrameworkExplainable2024}.
Closest to our setup, \citet{liCanMoranEigenvectors2025} run the same template of
synthetic SVCs and GeoShapley recovery across two geometries, but with an
analytic Moran-eigenvector basis. Whether GeoShapley recovers spatial
effects from learned, task-agnostic neural encoders that span planar and
spherical geometry has not been tested
\citep{russwurmGeographicLocationEncoding2024}. We fill that gap. We benchmark
eleven encoders from the TorchSpatial framework
\citep{wuTorchSpatialLocationEncoding2025} against a synthetic process with known
coefficients across three spatial scales, asking two questions: can GeoShapley
recover interpretable spatial effects from encoder-based models, and does the
choice of encoder architecture matter
\citep{raoMeasuringIntrinsicDimension2026}?

\section{Method}
\label{sec:method}

\begin{figure}[t]
  \centering
  \includegraphics[width=\columnwidth, trim=0 34 0 0, clip]%
    {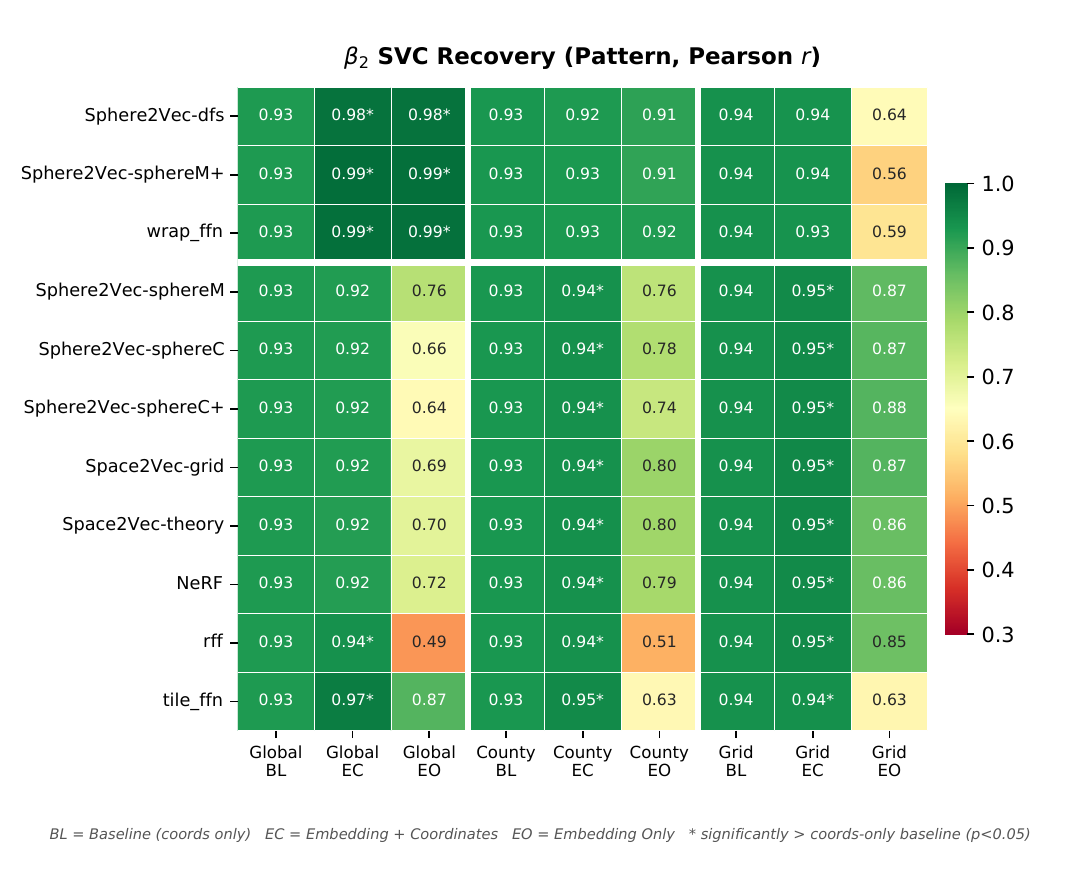}
  \Description{Heat map of Pearson correlation between GeoShapley-estimated and
  true secondary coefficients, with eleven location encoders as rows and nine
  columns given by three spatial scales crossed with three location-feature
  conditions.}
  \caption{$\beta_2$ recovery (Pearson $r$) per encoder across the three scales
  and the baseline (BL), embedding$+$coordinates (EC), and embedding-only (EO)
  conditions (MLP, untrained). $*$ marks values significantly above the
  coords-only baseline (paired Wilcoxon, $p<0.05$).
  The figure omits $\beta_1$, for which $r$ exceeded $0.98$ in nearly all conditions
  with raw coordinates (Table~\ref{tab:global}).}
  \label{fig:main_heatmap}
\end{figure}

\paragraph{GeoShapley.}
GeoShapley extends the classical Shapley value by grouping location features
into a single joint player ``GEO''. The intrinsic location effect is
\begin{equation}
  \phi_{\mathrm{GEO}}=\!\!\sum_{S\subseteq M\setminus\{\mathrm{GEO}\}}\!\!
  \frac{s!\,(p-s-g)!}{(p-g+1)!}\bigl(f(S\cup\{\mathrm{GEO}\})-f(S)\bigr),
\end{equation}
where $M$ is the feature set, $p$ the number of features, $s=|S|$, and $g$ the
number of grouped location features; spatially varying interaction effects
$\phi_{(\mathrm{GEO},j)}$ between GEO and a covariate $j$ yield the SVCs. The
GeoShapley package \citep{liGeoShapleyGameTheory2024} estimates these with
Kernel SHAP via weighted least squares \citep{covertImprovingKernelSHAPPractical2021}.

\paragraph{Data.}
Following \citet{liGeoShapleyGameTheory2024} and
\citet{fotheringhamMultiscaleGeographicallyWeighted2017}, the response is linear
in two non-spatial covariates with spatially varying coefficients,
\begin{equation}
  y_i=\beta_0(s_i)+\beta_1(s_i)X_{1i}+\beta_2(s_i)X_{2i}+\varepsilon_i,
  \quad \varepsilon_i\sim\mathcal{N}(0,\sigma^2),
  \label{eq:dgp}
\end{equation}
with $X_{1},X_{2}\sim\text{Uniform}(-2,2)$ and $\sigma=0.1$; we refer to
$\beta_1$ as the primary coefficient and $\beta_2$ as the secondary coefficient.
Three scales are
used: a $25\times25$ \emph{grid} ($N{=}625$); U.S.\ \emph{county} centroids
($N{\approx}3{,}000$); and a \emph{global} sample ($N{=}10{,}000$) over
$(-180,180)\times(-90,90)$. Each $N$ is set by its geometry -- the lattice, the
U.S.\ county set, and enough points to cover the globe -- rather than matched
across scales, so sample size, density, and sampling regularity co-vary with
scale. At grid/county scales $\beta_0$ is parabolic,
$\beta_1$ a linear gradient, and $\beta_2$ sinusoidal; at global scale the
surfaces are built on the sphere, with $\beta_2$ a multi-scale oscillation
continuous across the antimeridian. Data are split 80/20 train/test.

\paragraph{Encoders and training.}
Coordinates are transformed into a $D{=}8$ embedding by one of eleven
TorchSpatial encoders \citep{wuTorchSpatialLocationEncoding2025}:
Space2Vec-grid/-theory \citep{maiMultiScaleRepresentationLearning2020},
\texttt{tile\_ffn} \citep{tangImprovingImageClassification2015},%
\footnote{In the proceedings experiments, the TorchSpatial wrapper did not pass
the configured tile size to the lookup table. All locations therefore fell into
the same tile and received the same embedding, which carried no location
information. The results reported here use a tile side length equal to one
twentieth of the longer dimension of each study extent.}
\texttt{wrap\_ffn} \citep{aodhaPresenceOnlyGeographicalPriors2019}, NeRF
\citep{mildenhallNeRF2021}, five Sphere2Vec variants
\citep{maiSphere2VecGeneralpurposeLocation2023}, and random Fourier features
\citep{rahimiRandomFeaturesLargeScale2007}; a \texttt{none} baseline uses only
normalized coordinates. The feature vector concatenates the covariates, the $D$
embedding dimensions, and, in the default condition, normalized lon/lat. To
isolate the embedding from the coordinates available to every model, we also run
an \emph{embeddings-only} condition omitting lon/lat.
Two encoder conditions are compared: \emph{untrained} (weights frozen at random
init, testing architectural bias) and \emph{contrastively trained}, in which the
encoder is fit and then frozen by regressing pairwise embedding cosine similarity
onto a Gaussian kernel of the batch-normalized pairwise distance ($\sigma{=}0.25$;
Haversine at global scale, Euclidean otherwise) under an MSE loss (Adam,
$\eta{=}10^{-3}$, 500 epochs, batch $\min(512,N)$) -- a distance-preserving
objective rather than one built from sampled positive/negative pairs. Crossed with the
untrained/trained axis, this gives four configurations per scale and encoder. The downstream model
is an MLP (scikit-learn \citep{pedregosaScikitlearnMachineLearning2011}) with ReLU
activations trained by Adam, tuned per run with 5-fold randomized search (20
candidates) over hidden-layer sizes $\{(100,50),(150,100),(200,100)\}$, the
$\ell_2$ penalty $\alpha\in\{10^{-4},10^{-5},10^{-6}\}$, and the initial learning
rate $\in\{10^{-3},5{\times}10^{-4}\}$, with at most 5{,}000 iterations. Each
(scale, encoder, condition) is run for 25 seeds.
AI coding assistants (Claude Code) were used to write the experiment code, manage
cluster jobs, and assist the analysis; all experimental design, results, and
conclusions were verified by the authors.

\paragraph{Evaluation.}
Trained models are explained with GeoShapley over all $N$ observations, training
and held-out locations together, recovering
$\hat\beta_0,\hat\beta_1,\hat\beta_2$. The explainer uses a background set of up
to 200 subsampled training points and $n_{\mathrm{geo}}{=}D{+}2$ grouped location
features ($D$ where coordinates are omitted), and the resulting raw SVCs are
GWR-smoothed with adaptive bandwidth before evaluation. Pearson correlation
$r=\text{Pearson}(\hat\beta,\beta)$ against the ground truth is the primary
metric for spatial-pattern recovery. We average $r$ across the 25 seeds for
each condition. With raw coordinates, the standard deviation across seeds was
at most $0.02$ unless encoder training diverged. Without raw coordinates, it
reached $0.27$ for untrained encoders. Significance marks use unadjusted
$p$-values from paired Wilcoxon tests.

\section{Results}
\label{sec:results}

With raw coordinates, GeoShapley recovers $\beta_1$ well across encoders and
scales. Pearson $r$ exceeds $0.98$ in nearly all of these conditions
(Table~\ref{tab:global}). Recovery of $\beta_2$ varies more by encoder
(Figure~\ref{fig:main_heatmap}). At grid and county scales most
encoders reach $r\approx0.92$--$0.95$ for $\beta_2$ and the raw-coordinate
baseline is competitive with every embedding; the gains over baseline, though
occasionally significant, are negligible in magnitude ($\le 0.02$), so at these
scales any reasonable location representation supplies enough signal for the MLP.

At global scale the spread across encoders widens (Table~\ref{tab:global}).
Sphere2Vec-sphereM+ ($r{=}0.991$), \texttt{wrap\_ffn} ($r{=}0.988$), and
Sphere2Vec-dfs ($r{=}0.982$) score highest. The remaining encoders and the
baseline ($r{=}0.926$) range from $r{=}0.92$ to $0.97$. The three highest scorers
do not correspond to a single model family: \texttt{wrap\_ffn} is not a Sphere2Vec
model, and three other Sphere2Vec variants score no higher than the baseline.

\begin{figure*}[t]
  \centering
  \includegraphics[width=\textwidth]{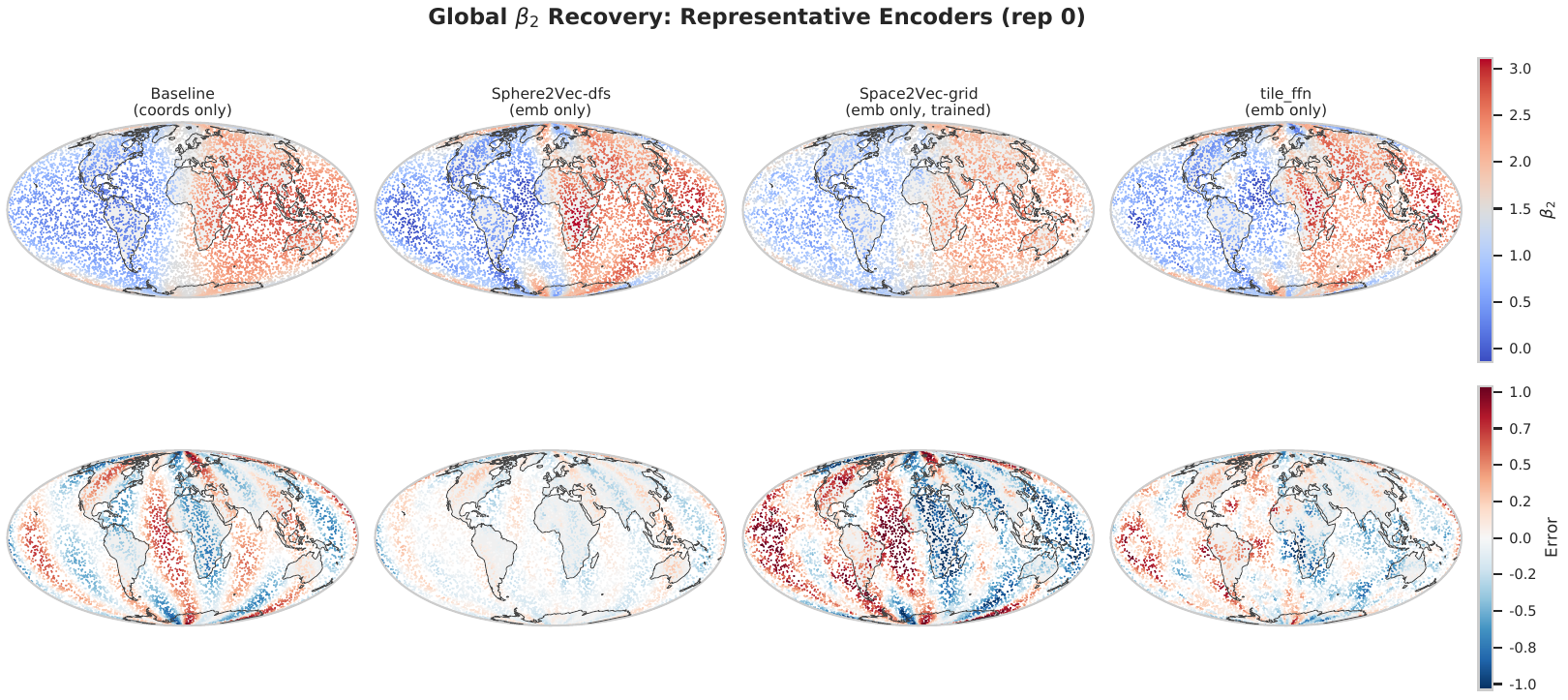}
  \Description{World maps for four location representations: the estimated
  secondary coefficient surface in the top row, and its error against the true
  surface in the bottom row.}
  \caption{Global $\beta_2$ estimates (top) and errors against the true surface
  (bottom) for representative encoders: the coordinates-only baseline, and
  Sphere2Vec-dfs, Space2Vec-grid (trained), and \texttt{tile\_ffn} in the
  embedding-only condition.}
  \label{fig:spatial_global}
\end{figure*}

Contrastive training generally does not improve, and often slightly degrades,
recovery: the $\Delta$ column in Table~\ref{tab:global} is negative for most
encoders, indicating that the frozen random initialization already provides
sufficient positional signal. The clearest
exception is Sphere2Vec-sphereM, where training improves global $\beta_2$ by
$0.030$; training was occasionally
unstable (e.g.\ \texttt{wrap\_ffn} at grid scale). The baseline stays competitive
throughout: embeddings do not automatically improve SVC recovery over plain
normalized coordinates, echoing \citet{liCanMoranEigenvectors2025}, who likewise
found coordinate-only models hard to beat with an analytic eigenvector basis. The
highest-scoring global encoders are the exception, pulling ahead of the baseline.

\begin{table}[tb]
\centering
\caption{Pearson correlation ($r$) by encoder at global scale, untrained vs.\
contrastively trained ($\Delta = $ trained $-$ untrained). Bold marks the three highest
$\beta_2$ recovery values in each condition.}
\label{tab:global}
\footnotesize
\setlength{\tabcolsep}{4pt}
\begin{tabular}{lccc|ccc}
\toprule
& \multicolumn{3}{c|}{$\beta_1$} & \multicolumn{3}{c}{$\beta_2$} \\
\cmidrule(lr){2-4} \cmidrule(lr){5-7}
Encoder & Untr. & Tr. & $\Delta$ & Untr. & Tr. & $\Delta$ \\
\midrule
NeRF                & 0.996 & 0.996 & $-$.000 & 0.919 & 0.897 & $-$.022 \\
Space2Vec-grid      & 0.996 & 0.996 & $-$.001 & 0.921 & 0.918 & $-$.004 \\
Space2Vec-theory    & 0.996 & 0.995 & $-$.001 & 0.920 & 0.914 & $-$.006 \\
Sphere2Vec-dfs      & 0.999 & 0.999 & $-$.000 & \textbf{0.982} & \textbf{0.979} & $-$.003 \\
Sphere2Vec-sphereC  & 0.997 & 0.995 & $-$.001 & 0.921 & 0.912 & $-$.009 \\
Sphere2Vec-sphereC+ & 0.996 & 0.994 & $-$.002 & 0.922 & 0.922 & $+$.000 \\
Sphere2Vec-sphereM  & 0.997 & 0.996 & $-$.000 & 0.922 & 0.952 & $+$.030 \\
Sphere2Vec-sphereM+ & 0.999 & 0.999 & $-$.000 & \textbf{0.991} & \textbf{0.984} & $-$.007 \\
none (baseline)     & 0.998 & 0.998 & $\;\;$.000 & 0.926 & 0.926 & $\;\;$.000 \\
rff                 & 0.995 & 0.994 & $-$.002 & 0.938 & 0.928 & $-$.010 \\
\texttt{tile\_ffn}  & 0.997 & 0.997 & $-$.001 & 0.967 & 0.948 & $-$.020 \\
\texttt{wrap\_ffn}  & 0.999 & 0.997 & $-$.002 & \textbf{0.988} & \textbf{0.989} & $+$.001 \\
\bottomrule
\end{tabular}
\end{table}

\paragraph{Embeddings without coordinates.} Because lon/lat are present in every
condition above, that comparison captures only the embedding's small
\emph{marginal} value over raw coordinates. Removing them makes the embedding the
sole location representation (the EO columns of
Figure~\ref{fig:main_heatmap}), and large architectural differences emerge that
invert with scale. At global scale Sphere2Vec-sphereM+ ($r{=}0.988$),
\texttt{wrap\_ffn} ($0.985$), and Sphere2Vec-dfs ($0.984$) retain essentially all
of their $\beta_2$ recovery and \texttt{tile\_ffn} retains most of it ($0.875$),
while the others drop sharply (NeRF $0.715$, Space2Vec-grid $0.689$, \texttt{rff}
$0.485$; Figure~\ref{fig:spatial_global}). At grid scale the ordering \emph{reverses}: most encoders reach
$r\approx0.85$--$0.88$ --- NeRF ($0.856$), Space2Vec-grid ($0.875$), and
\texttt{rff} ($0.852$) among them --- while those three global leaders fall
to $r\approx0.56$--$0.64$. Without
coordinates, then, the encoders that recover global structure are largely the
ones that recover the local grid least well, and vice versa.

\paragraph{Robustness to nonlinearity.} Because Eq.~\eqref{eq:dgp} is linear in the
covariates, we repeated all experiments with a nonlinear process that adds a
global quadratic term and offset, $y_i = \beta_0(s_i) + [\beta_1(s_i)X_{1i} +
X_{1i}^2] + [\beta_2(s_i)X_{2i} + 2X_{2i}] + \varepsilon_i$. Recovery of
$\beta_2$ was essentially unchanged at every scale (mean $|\Delta r|\le0.005$),
and the global-scale ordering was preserved (Sphere2Vec-sphereM+ $0.990$,
\texttt{wrap\_ffn} $0.984$, Sphere2Vec-dfs $0.979$ vs.\ baseline $0.916$). The
gap is wider on amplitude than on correlation: those three recover $93$--$96\%$
of the true $\beta_2$ range (OLS slope of estimated on true) against
$76$--$87\%$ for the remaining encoders and the baseline, so their advantage is
one of magnitude, not only of pattern, and is not an artifact of a linear
data-generating process. Recovery of $\beta_1$ degraded modestly ($\Delta r$ from
$-0.02$ to $-0.07$) on the coefficient sharing a feature with the quadratic
term, which GeoShapley absorbs into that main effect.

\section{Discussion and Conclusion}
\label{sec:discussion}

Encoder architecture matters in a scale-dependent way. At grid and county scales,
when raw coordinates are available, differences between encoders are small and
normalized coordinates suffice; at
global scale, where $\beta_2$ varies across the $\pm180^{\circ}$ antimeridian,
they become substantial. This extends intrinsic-dimensionality analyses of the same
encoder zoo \citep{raoMeasuringIntrinsicDimension2026} and GeoShapley benchmarks
against GWR/MGWR \citep{liuEnsembleFrameworkExplainable2024} to learned
high-dimensional encoders. Untrained encoders generally matched or outperformed
their contrastively trained counterparts in this benchmark.

Limitations: the data-generating process is synthetic and smooth, so the
encoders that performed best here need not suit sharp or discontinuous effects;
sample size and sampling geometry co-vary with scale, so scale is not isolated
from $N$; explanations pool training and held-out locations rather than
separating them; correlation is the headline metric, and although the amplitude
result above is magnitude-sensitive we do not tabulate absolute-error metrics;
the embedding dimension is fixed at $D{=}8$ for tractability; and results are
specific to the MLP and explainer settings.

Overall, GeoShapley drops into a
location-encoder pipeline as is, without an interpretable embedding space, and
reliably recovers spatially structured effects. Practically, encoder choice
matters most at global scale, contrastive pre-training should not be assumed to
help, and the embeddings-only reversal argues for keeping raw coordinates
alongside any embedding so that recovery degrades gracefully when the embedding is
a poor match for the scale. Auditing whether a model has learned
spatial effects is a step toward spatial-fairness frameworks
\citep{saxenaSpatialFairnessCase2024}. Future work should
sweep the embedding dimension, vary $N$ within each geometry to separate scale
from sample size, and test better-aligned contrastive objectives, discontinuous
spatial-effect regimes, and real data with partially known spatial processes.

\paragraph{Code and data availability.} Code to reproduce all experiments,
including the synthetic data generators, encoder wrappers, and GeoShapley
evaluation, is at \url{https://github.com/cybergis/loc-enc-svcs}.

\begin{acks}
This project was supported by Award No.~15PNIJ-24-GG-01573-RESS, awarded by the
National Institute of Justice, Office of Justice Programs, U.S.\ Department of
Justice. The opinions, findings, and conclusions or recommendations expressed in
this publication are those of the author(s) and do not necessarily reflect those
of the Department of Justice.
This research used the TGI RAILS compute resource (National Science Foundation
(NSF) award number: OAC-2232860, Taylor Geospatial Institute) and Virtual ROGER,
supported by the CyberGIS Center for
Advanced Digital and Spatial Studies and the School of Earth, Society and
Environment at the University of Illinois Urbana-Champaign.
This research was also supported by I-GUIDE, an Institute funded by NSF, under
award number 2118329.
\end{acks}

\bibliographystyle{ACM-Reference-Format}
\bibliography{references}

\end{document}